\documentclass[10pt,twocolumn,letterpaper]{article}

\usepackage{cvpr}              

\usepackage{graphicx}
\usepackage{amsmath}
\usepackage{amssymb}
\usepackage{booktabs}
\usepackage{diagbox}
\usepackage{multirow}
\usepackage{float}

\usepackage[pagebackref,breaklinks,colorlinks]{hyperref}

\usepackage[capitalize]{cleveref}
\crefname{section}{Sec.}{Secs.}
\Crefname{section}{Section}{Sections}
\Crefname{table}{Table}{Tables}
\crefname{table}{Tab.}{Tabs.}

\def\cvprPaperID{13211} 
\def\confName{CVPR}
\def\confYear{2025}

\begin{document}

\title{HSI: A Holistic Style Injector for Arbitrary Style Transfer}

\author{Shuhao Zhang, Hui Kang, Yang Liu, Fang Mei, Hongjuan Li\\
Jilin University\\
\shortstack{\tt\small shzhang22@mails.jlu.edu.cn, kanghui@jlu.edu.cn, \\ \tt\small liuy24@mails.jlu.edu.cn, meifang@jlu.edu.cn, hongjuan23@mails.jlu.edu.cn}
}
\maketitle


\begin{figure*}[t]
\centering
\setlength{\abovecaptionskip}{-0.1cm}
\includegraphics[width=6.0in, height=2.6in]{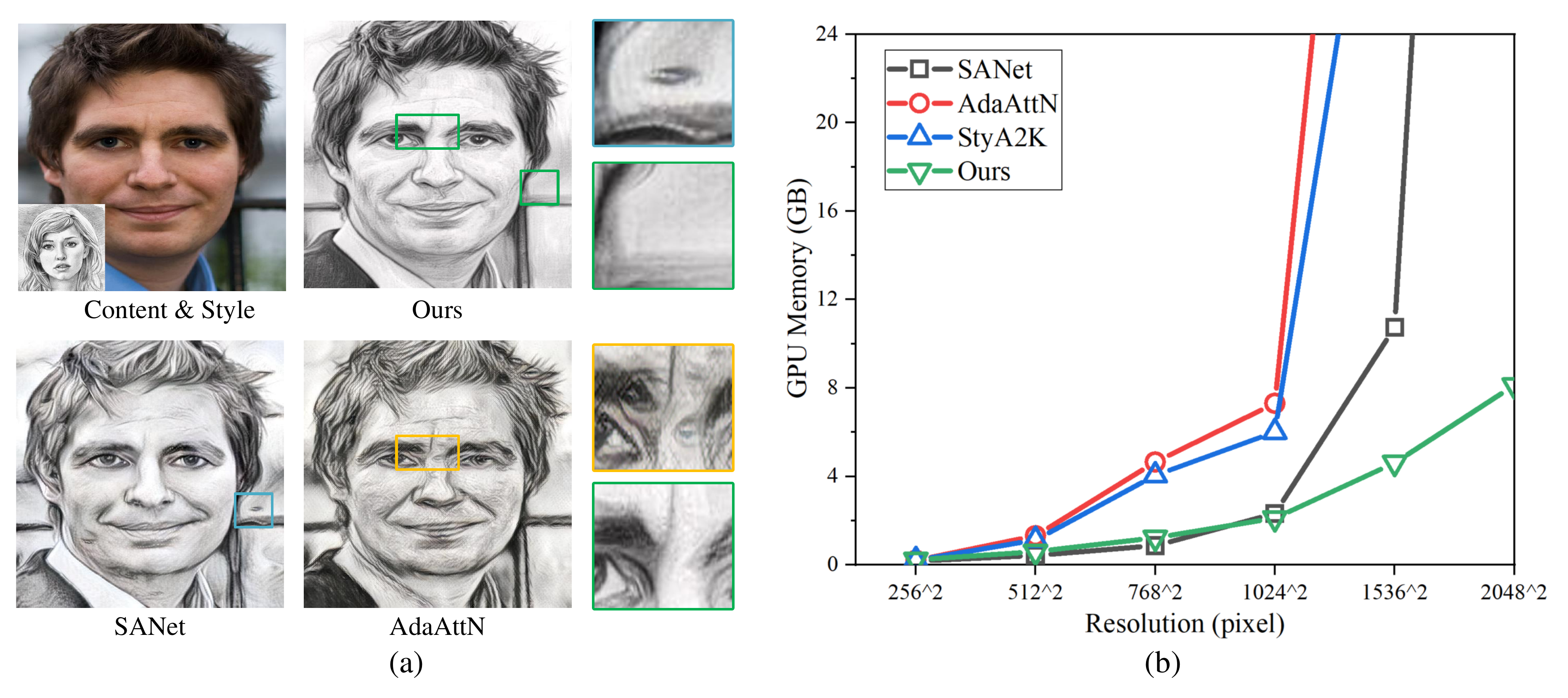}
\caption{The comparison between our method and some attention-based methods. (a) Image stylization results. Compared to SANet \cite{sanet} and AdaAttN \cite{adaattn}, the generated image of our method is more consistent with original content and more harmonious in style patterns. (b) GPU memory consumption in different resolutions. Compared to other methods,  our method successfully renders images from 256 $\times$ 256 to 2048 $\times$ 2048 resolution on a 24GB GPU (4090Ti) without running out of memory.}
\label{motivation}
\vspace{-0.5cm}
\end{figure*}

\begin{abstract}
Attention-based arbitrary style transfer methods have gained significant attention recently due to their impressive ability to synthesize style details. However, the point-wise matching within the attention mechanism may overly focus on local patterns such that neglect the remarkable global features of style images. Additionally, when processing large images, the quadratic complexity of the attention mechanism will bring high computational load.
To alleviate above problems, we propose Holistic Style Injector (HSI), a novel attention-style transformation module to deliver artistic expression of target style. 
Specifically, HSI performs stylization only based on global style representation that is more in line with the characteristics of style transfer, to avoid generating local disharmonious patterns in stylized images. 
Moreover, we propose a dual relation learning mechanism inside the HSI to dynamically render images by leveraging semantic similarity in content and style,  ensuring the stylized images preserve the original content and improve style fidelity.
Note that the proposed HSI achieves linear computational complexity because it establishes feature mapping through element-wise multiplication rather than matrix multiplication.
Qualitative and quantitative results demonstrate that our method outperforms state-of-the-art approaches in both effectiveness and efficiency.

\end{abstract}

\section{Introduction}
\label{sec:intro}

Arbitrary Style Transfer (AST) is an artwork creating technique to transform a natural image (i.e., content image) with specific artistic semantic by learning the representative style elements of the artistic images (i.e., style image). 
Since Gates et al.\cite{gatys}  proposed the first neural style transfer algorithm that integrates the Gram statistics of style image into content image, previous methods have demonstrated impressive stylization results by adopting global feature distribution alignment methods \cite{adain,wct,efdm} and local patch matching strategies \cite{chenqianqi,Avatar-net,graph_cut}.  However, these methods can not capture the semantic correspondences between content and style inputs and usually merge style elements with content features directly, thus leading to distorted regions or texture patterns.

To address these problems, researchers have proposed attention-based AST methods\cite{adaattn,sanet,stytr2,s2wat}, to establish the point-to-point semantic correlation between content and style features by equipping the attention module \cite{transformer}.
Due to the adaptive matching of content features and style features, attention mechanism enabled methods thus demonstrating impressive stylization performance, whereas there are still existing three notable limitations:  
(1) It is intuitively challenging to deliver representative style patterns only with a single-point feature. The attention mechanism establishes the semantic relations between content and style features at a point-wise level, the holistic style representations may be neglected. 
(2) The exponential calculation of the softmax function may produce biased patterns in the resulting images. Specifically, the attention module may overly concentrate on a prominent style area and ignore the holistic distribution,  which may lead to disharmonious patterns.
As shown in Figure \ref{motivation}(a), the eye pattern of the style image appears in multiple regions in the stylized image for SANet \cite{sanet} and AdaAttN \cite{adaattn}.
(3) The matrix multiplication will cause quadratic complexity.
As shown in Figure \ref{motivation}(b), with the growth of the image resolution, the memory consumption of the attention-based approaches (SANet \cite{sanet}, AdaAttN \cite{adaattn} and StyA2K \cite{stya2k}) increases sharply and ultimately cause GPU memory overflow (24GB on 4090Ti).
Therefore, the original attention module for the AST task is incompetent to learn the holistic style representations and is inefficient for end-to-end style transfer based on the above drawbacks.

A new all-to-key attention named StyA2K \cite{stya2k} is proposed to solve these problems recently. It implements a patch-wise aggregation strategy that uses a block region of features instead of a point feature to represent style patterns.
While StyA2K puts more style elements to interact with the content feature, it is still challenging in conveying a cohesive style pattern, especially for the artistic image that prioritize style integrity.
Moreover, although the patch-wise calculation strategy has reduced the computational load, its complexity still remains at a quadratic level.

Based on the analysis of existing works, it can be found that it is hard to convey the entire style pattern using a single-point feature. Therefore, by extracting the global and local style information simultaneously, the style representation will be established with a more stronger semantic association. Based on the motivations above, we propose a light transfer module called the Holistic Style Injector (HSI) to achieve effective and efficient style transfer. 
The proposed HSI has three traits: 
(1) Global styles extraction: Instead of calculating point-to-point similarity to guide style transfer, HSI directly extracts a range of global style statistical features and uses them to establish a straight connection with content features. This operation can render sufficiently comprehensive style characteristics onto the content while avoiding style bias caused by focusing on local areas overly.
(2) Dynamic dual relations construction: HSI constructs local-content-to-global-style and global-content-to-global-style relations simultaneously to improve the stylization quality. When the content image is semantically close with the style image, the global-content-to-global-style relation will strengthen, enhancing the overall harmony of the stylized image.
(3) Linear-complexity transfer process: HSI employs element-wise multiplication to establish content-style relations, which can perform the transfer process in linear complexity. This multiplication dramatically reduces the computational complexity compared with the matrix multiplication in the attention mechanism. 
To our knowledge, this is the first study that successfully exploits the power of element-wise multiplication in the style transfer.
In summary, the main contributions of this paper are listed as follows:
\begin{itemize}
    \item This is the first work to propose element-wise multiplication as a fundamental component for building semantic relations between content and style features in arbitrary style transfer. Based on this, we introduce a novel style synthesis module named HSI that is structurally similar to self-attention, yet it produces higher-quality stylized images only with linear computation processes.

    \item We propose to incorporate multiple types of style statistical features into content features, and establish the dynamical relationship of local-content-to-global-style and global-content-to-global style. These strategies enrich the diversity of stylistic elements in stylized images and improve their style fidelity.

    \item Extensive experiments demonstrate the excellent performance of our method in effectiveness and efficiency for AST task.
\end{itemize}

\section{Related Work}
\label{sec:relatedWork}
\subsection{Arbitrary Style Transfer}
Since Gaty et al. \cite{gatys} introduced the first neural style transfer model using pre-trained neural networks to synthesize artistic styles through iterative optimization, a variety of arbitrary style transfer algorithms \cite{artflow,cl1,IEC,styleformer,Avatar-net,deep_reshufffle} have been developed. Generally, AST methods can be divided into two categories: Global transformation based methods \cite{adain,wct,gt_ast1,gt_ast2,efdm} and local matching based methods \cite{Avatar-net,chenqianqi,sanet,aespa-net,stytr2}. 
For the former, their common idea is to directly match the style statistical distribution at the global feature level. 
Typical examples are AdaIN \cite{adain} and WCT \cite{wct}. 
Specifically, AdaIN directly matches the mean and variance between content and style features at the channel level. Similarly, WCT introduces whitening and coloring operations to align the overall style distribution. While these methods are efficient for stylization, they often struggle with visual quality because they can not capture the semantic relationship  content and style elements. 

For the latter, they emphasize the local semantic alignment between content features and style features. StyleSwap \cite{chenqianqi} and Avatar-Net \cite{Avatar-net} are early representatives of this class of methods. 
They fuse style elements by calculating the similarity between content features and style features on a patch-wise level, often achieving superior visual quality compared to the global transformation based methods. 
With self-attention \cite{transformer} gaining popularity across various fields,  many researchers began exploring attention-based AST methods \cite{adaattn,sanet,IEC}. 
These methods can render the content image according to the point-wise semantic similarity learned adaptively between content features and style features. 
SANet \cite{sanet} directly uses the self-attention module to perform style transfer in the feature space. Based on SANet, AdaAttN \cite{adaattn} further integrates the advantages of AdaIN algorithm for feature normalization. Furthermore, StyTr$^2$ \cite{stytr2} and S2WAT \cite{s2wat} employ transformer components \cite{vision_transformer} to extract style features and transfer styles. 
While attention-based methods can effectively preserve semantic information of content and style, they often overly focus on aggregating local patterns. This may lead to content distortion of the generated images due to the repetitive and scattered style patterns.
In addition, the quadratic complexity is a common challenge for the attention-based methods. 
Recently, StyA2K \cite{stya2k} try to address above problems by proposing an all-to-key attention module to improve style transfer stability and efficiency. Compared with the point-wise similarity calculation in self-attention, all-to-key attention builds similarity map in a patch-wise feature level. 
However, the generated results are often under-stylized due to the excessive emphasis on structural stability. 
Moreover, its computational complexity remains quadratic with the image size despite the improvement in efficiency. 
In this paper, we aim to explore a more effective transformation module that achieves high-quality stylization with sufficiently low computational complexity.

\begin{figure} [t!]
\centering
\setlength{\abovecaptionskip}{-0.3cm}
\includegraphics[width=3.4in, height=1.2in]{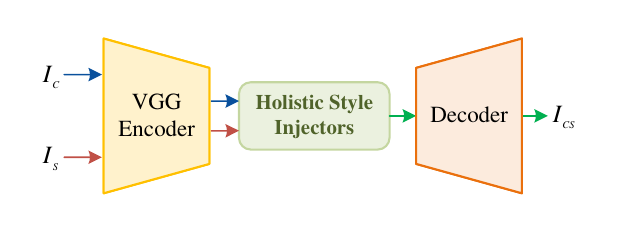}
\caption{The network framework of our method.}
\label{framework}
\vspace{-0.4cm}
\end{figure}

\subsection{Attention Mechanism}
The attention mechanism was first introduced in machine translation by Bahdanau et al \cite{att2015}. The most well-known variant is self-attention, which was introduced in the Transformer model \cite{transformer}. Self-attention is widely used in natural language processing \cite{nlp_att1,nlp_att2,nlp_att3,nlp_att4} and computer vision \cite{cv_att1,cv_att2,cv_att3,cv_att4} due to its ability to establish long-distance dependencies. 
Its robust scalability and adaptability in establishing similarity relationships between local features have attracted many researchers to apply it for improving performance in their respective fields.
However, in image generation tasks like AST \cite{sanet,adaattn}, 
the original dense relational learning mechanism in self-attention may negatively affect visual quality due to focusing on local style areas.
Moreover, the high computational complexity of self-attention is also a common challenge. 
In this paper, we seriously consider these issues and explore how style patterns can better match the content during the feature interaction, and propose an effective and efficient style transformation module for AST task.

\begin{figure*} [t!]
\centering
\setlength{\abovecaptionskip}{-0.3cm}
\includegraphics[width=5.4in, height=4.0in]{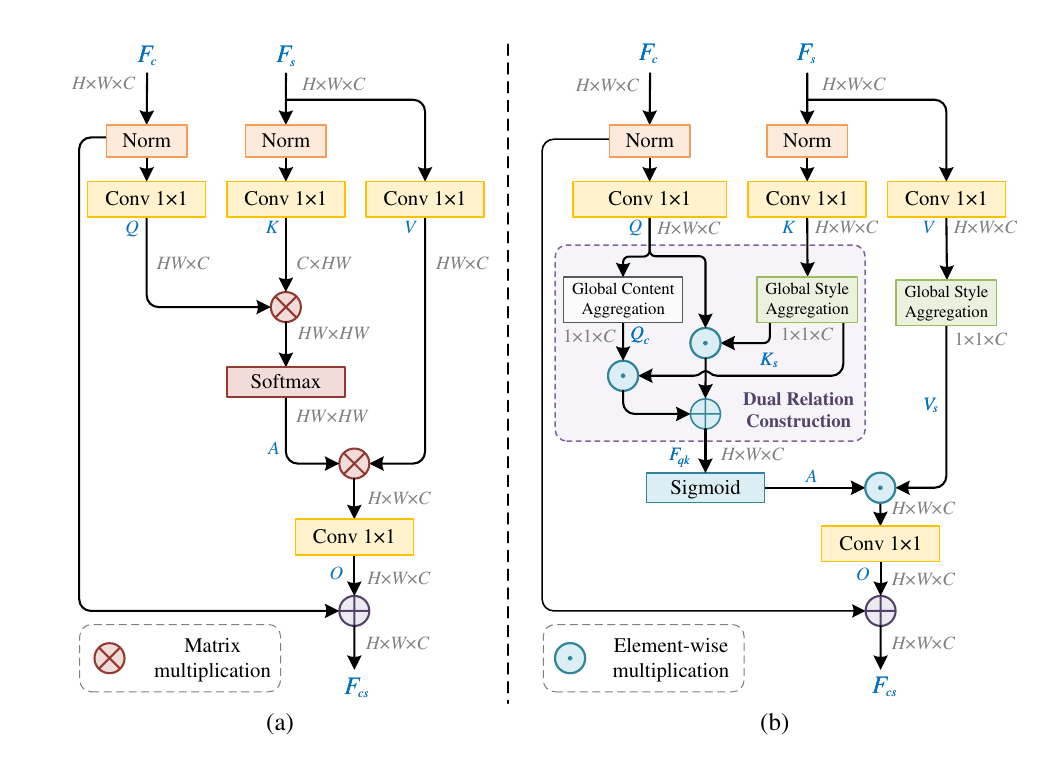}
\caption{The structure and feature encoding process comparison with self-attention (a) and our HSI module (b). HSI has a similar structure to self-attention, which uses element-wise multiplication instead of matrix multiplication to model the semantic similarity of content features and style features.}
\label{self_att}
\vspace{-0.5cm}
\end{figure*}

\section{Proposed Method}
\label{sec:method}
\subsection{Overall Framework}
The overall style transfer network is shown in Figure \ref{framework}. A simple encoder-decoder architecture is adopted in this paper, where encoder $E(\cdot)$ is a parameter-fixed pre-trained VGG-19 \cite{vgg} network  (up to relu\_4\_1 layer). After extracting the content feature $F_c$ and style feature $F_s$, a chain of connected HSI modules will receive both features to complete style transform in feature space, which generate the synthetic feature $F_{cs}$:
\begin{equation}
  F_{cs} = \text{HSIs}(E(I_c), E(I_s))
  \label{eq:hsi}
\end{equation}

Then, the decoder will invert the $F_{cs}$ to the image space that produces the stylized image $I_{cs}$:
\begin{equation}
  I_{cs} = D(F_{cs})
  \label{eq:decoder}
\end{equation}
where the structure of decoder is an mirror of the encoder. Following \cite{wct}, to alleviate the checkerboard effects, we adopt the nearest up-sampling to replace the pooling layer. Moreover, the reflection padding is used in encoder and decoder to prevent the board collapse.

\subsection{Revisit Attention Mechanism in AST}
In AST, attention mechanism refers to the self-attention\cite{transformer} by default. Benefiting from flexible relation building ability, attention module can adaptively embed style patterns into content features of each position for stylization. Figure \ref{self_att}(a) presents the overall encoding process.  Query ($Q$), key ($K$) and value ($V$) are the core encoding feature of attention module, they can be obtained by:
\begin{equation}
  Q = f_q(Norm(F_c)), K = f_k(Norm(F_s)), V = f_v(F_s)
  \label{eq:qkv}
\end{equation}
where $f_q$, $f_k$ and $f_v$ denote the convolution layer which kernel usually with size of $1 \times 1$. $Norm(\cdot)$ is the mean-variance channel-wise normalization. We can obtain the attention score $A$ by
\begin{equation}
  A = softmax(Q \otimes K^T)
  \label{eq:attention}
\end{equation}
where $\otimes$ is the matrix multiplication. Attention map $A$ indicates the point-to-point similarity score between content feature $F_c$ and style feature $F_s$. By the similarity map $A$, the extracted style elements can be transferred on the content features through
\begin{equation}
  O = f_o(A \otimes V)
  \label{eq:output_att}
\end{equation}
where $f_o$ denotes the convolution layer like $f_q$, $f_k$ and $f_v$. $O$ is the synthetic feature for generating stylized image, which is also added to the input content feature $F_c$ to improve the structure consistency, as shown in Figure \ref{self_att}(a).


As analyzed in the Introduction, although the attention mechanism establishes a point-to-point matching relationship between content features and style features to improve semantic consistency in stylization results, it may overfocus on a particular style region, leading to an inharmonious pattern in the resulting image. In addition, the attention mechanism has quadratic computational complexity with the size of the input image. The detailed complexity analysis is as follows.

\textbf{Complexity Analysis.} We assume $Q$, $K$ and $V$ with size of ($H$, $W$, $C$). The computation of self-attention mainly focuses on two phases. One is the dot product in each position between $Q$ and $K$ for generating attention map $A$ that also with size of ($H \times W$, $H \times W$). Another is the weighted summation of the $A$ and $V$ in Equation \ref{eq:output_att}. The computational complexity of both are  $\mathcal{O}$(($H \times W$)$^2 \times C$), so the computational complexity of self-attention is $\mathcal{O}((H \times W)^2 \times C)$.


\subsection{Holistic Style Injector}
Although the attention mechanism can produce promising stylization results in AST, it also has obvious limitations in terms of style rendering and computational efficiency.
To overcome these issues, we propose a transformation module called the Holistic Style Injector (HSI), which builds more comprehensive correlations between style and content features for improved stylization while reducing computational complexity.
On the whole, HSI and the attention module have similar processing processes, as shown in Figure \ref{self_att}. 
First, the content feature $F_c$ and the style feature $F_s$ are normalized and linearly processed to obtain the key triplet ($Q$, $K$, $V$) for stylization. Next, $Q$ will interact with $K$ and their results further be normalized probabilistically to generate the attention score $A$ that hides the semantic relationships between the content and style features. Then, the style representation $V$ is weighted and summed with $A$, followed by a linear transform to synthesize style elements and produce the output feature $O$. Finally, a residual connection is established to link the input content feature $F_c$ with $O$, ensuring a more consistent content structure in the stylization results.

There are noticeable differences between HSI and attention module in constructing relationships between content and style features. In contrast with attention module, HSI introduces three key innovations: (1) Global styles extraction, (2) Dynamic dual relations construction, and (3) Linear-complexity transfer process.

\begin{figure} [t!]
\vspace{-0.4cm}
\centering
\setlength{\abovecaptionskip}{-0.3cm}
\includegraphics[width=3.6in, height=2.1in]{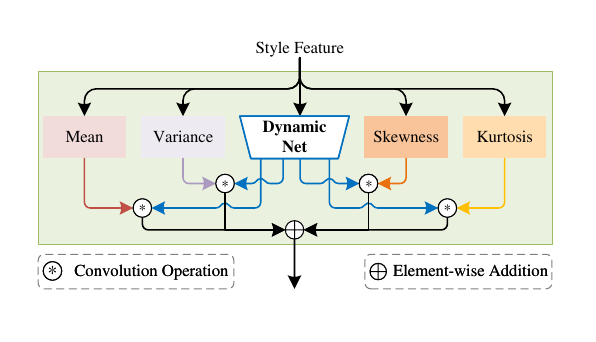}
\caption{The detailed illustration of global style aggregation in solid green box of Figure \ref{self_att}(b).}
\label{global_style_agg}
\vspace{-0.5cm}
\end{figure}

\textbf{Global Styles Extraction.}
Unlike the attention module that calculates the point-wise feature similarity, HSI focuses on mining the relationship between global style and content features. To describe the style pattern more comprehensively, we utilize four channel-wise statistical features, including mean, variance, skewness and kurtosis, combining the content features perform the stylization, as shown in Figure \ref{global_style_agg}. Specifically, the mean and variance express the overall tone and variation degree of style elements, while skewness and kurtosis indicate the symmetry and concentration of the distribution of style features. Their definitions are as follows:
\begin{align}
    \mu &= \frac{1}{H \times W} \sum_{h=1}^{H} \sum_{w=1}^{W} x_{h, w, c} \notag \\
    \sigma^2 &= \frac{1}{H \times W} \sum_{h=1}^{H} \sum_{w=1}^{W} (x_{h, w, c} - \mu)^2  \\
    \gamma_1 &= \frac{1}{H \times W} \sum_{h=1}^{H} \sum_{w=1}^{W} \left(\frac{x_{h, w, c} - \mu}{\sigma}\right)^3 \notag \\
    \gamma_2 &= \frac{1}{H \times W} \sum_{h=1}^{H} \sum_{w=1}^{W} \left(\frac{x_{h, w, c} - \mu}{\sigma}\right)^4 \notag
\end{align}
where $\mu$, $\sigma$, $\gamma_1$ and $\gamma_2$ denote the channel-wise mean, standard deviation, skewness and kurtosis respectively. Then, we employ an dynamic network \cite{dynamic_net} that transforms average and maximum value of style component $K$ into the weight $W$ and bias $b$:
\begin{align}
    W &= AvgPool(Conv(K)) \notag \\
    b &= MaxPool(Conv(K)) 
\end{align}
where $Conv(\cdot)$ denotes the depth-wise separable convolution, $AvgPool(\cdot)$ and $MaxPool(\cdot)$ are global average/max pooling operation. 
Note that each statistical feature has its own $W$ and $b$. Then, four style statistical features are combined in a weighted summation manner and the result will interact with the content feature.

\textbf{Dynamic Dual Relations Construction.} 
To better exploit the style patterns of the original image, we establish two relationships: local-content-to-global-style and global-content-to-global-style, as illustrated in the purple dashed box of Figure \ref{self_att}(b). 
Specifically, the global style feature $K_s$ interacts with $Q$, which contains refined content information, and the global content feature $Q_c$, which encompasses overall information from the $Q$. This interaction process is achieved through:
\begin{align}
     F_{qk} = \lambda_g \times (Q_c \odot K_s) \oplus (1 - \lambda_g) \times (Q \odot K_s)
\end{align}
where $\odot$ and $\oplus$ denote the element-wise multiplication and element-wise addition with broadcast mechanism. The fused feature $F_{qk}$ is obtained based on the similarity coefficient $\lambda_g$:

\begin{align}
     \lambda_g = \frac{\frac{Q \cdot K}{\|Q\|\|K\|} + 1}{2}
\end{align}
In our idea, we pay more attention to the global semantic relation between content and style feature. Therefore, when both the content image and style image are semantically similar, such as both being human faces, the value of $\lambda_g$ will be larger. 
This emphasizes the importance of the global-content-to-global-style relation for stylization, which can help reduce the disharmonious patterns caused by excessively focusing on local style patterns.

\textbf{Linear-Complexity Transfer Process.}
The linear complexity of HSI is mainly reflected on the element-wise multiplication operation, which  appears in several nodes of the HSI module, as shown in Figure \ref{self_att}(b). When the size of input feature are ($H$, $W$, $C$), the complexity of element-wise multiplication is $\mathcal{O}(H \times W \times C)$ because only the element multiplication at corresponding positions is involved. Therefore, the complexity of HSI module is $\mathcal{O}(H \times W \times C)$.

\begin{figure*} [t!]
\centering
\setlength{\abovecaptionskip}{-0.1cm}
\includegraphics[width=7.0in, height=3.0in]{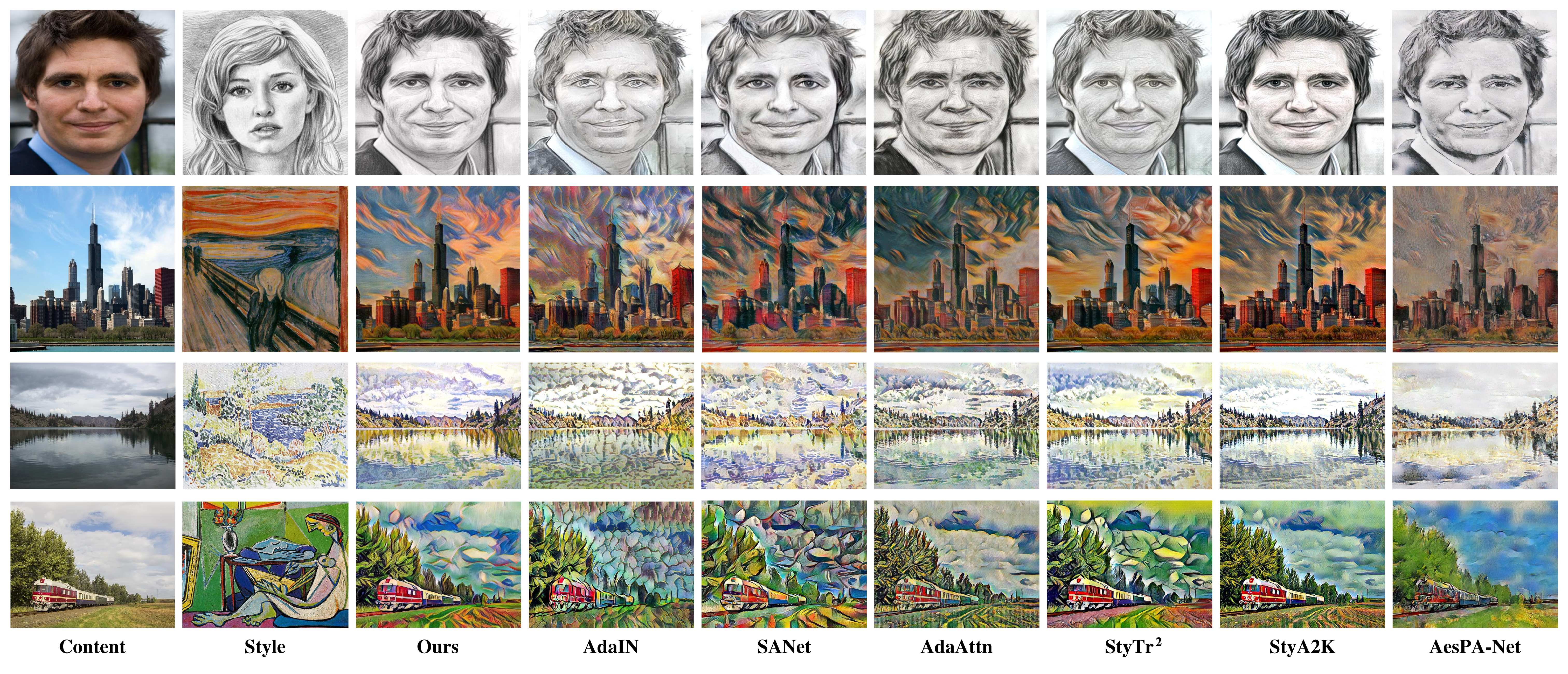}
\caption{Qualitative comparisons with state-of-the-art AST methods. Zoom in for a better view.}
\label{exp_results}
\vspace{-0.5cm}
\end{figure*}

\subsection{Loss Functions}
We optimize the entire network using three term of loss functions. The first term is the style loss $\mathcal{L}_s$ as defined in \cite{adain}, which is responsible for aligning the mean $\mu$ and standard deviation $\sigma$ between the stylized image $I_{cs}$ and the style image $I_s$ in VGG feature space:
\begin{equation}
\begin{split}
  \mathcal{L}_s = \sum^{N_l}_{i=1}\|\mu(\phi_{i}(I_{cs})) - \mu(\phi_{i}(I_{s}))\|_{2} \\ + \|\sigma(\phi_{i}(I_{cs})) - \sigma(\phi_{i}(I_{s}))\|_{2}
  \label{eq:ls}
  \end{split}    
\end{equation}
where $\phi_{i}$ denotes $l$th layer of the pre-trained VGG network and $N_l$ represents the number of layers.
The second term is the content loss $\mathcal{L}_c$ for preserving content structure:
\begin{equation}
  \mathcal{L}_c = \sum^{N_l}_{i=1}\|\phi_{i}(I_{cs}) - \phi_{i}(I_{c})\|_{2} + \|\phi_{i}(I_{cs}^{gray}) - \phi_{i}(I_{c}^{gray})\|_{2}
  \label{eq:lc}   
\end{equation}
where $I_{*}^{gray}$ represents the corresponding gray image. In addition, we adopt a adversarial loss $\mathcal{L}_{adv}$ based on a multi-scale discriminator to improve holistic stylization effect. 

The three losses are summed to consist the full loss:
\begin{equation}
  \mathcal{L} = \lambda_s \mathcal{L}_{s} + \lambda_c \mathcal{L}_{c} + {\lambda}_{adv} \mathcal{L}_{adv}
  \label{eq:loss}   
\end{equation}
where $\lambda_s$, $\lambda_c$ and $\lambda_{adv}$ are the weighting factors. We set $\lambda_s$, $\lambda_c$ and $\lambda_{adv}$ as 60, 5 and 50 to strike a balance among losses.

\begin{table*}[]
\centering
\small
\caption{Quantitative comparisons with state-of-the-art AST methods. We employ average content loss, style loss, LPIPS and FID metrics to measure the visual quality of the generated images.}
\renewcommand{\arraystretch}{1.0}
\setlength{\tabcolsep}{3pt}
\label{fid}
\begin{tabular}{@{}lccccccc@{}}
\toprule
\textbf{Method}             & \textbf{AdaIN} \cite{adain} & \textbf{SANet} \cite{sanet} & \textbf{AdaAttN} \cite{adaattn} & \textbf{StyTr$^2$} \cite{stytr2} & \textbf{StyA2K} \cite{stya2k}       & \textbf{AesPA-Net} \cite{aespa-net} & \textbf{Ours}           \\ \midrule
Content Loss $\downarrow$ & 0.97  & 1.18  & 1.21    & 0.69   & \textbf{0.59} & 0.63      & 0.62           \\
Style Loss $\downarrow$ & 1.44  & 1.26  & 1.52    & 1.34   & 1.21          & 0.99      & \textbf{0.95}  \\
LPIPS $\downarrow$ & 0.65  & 0.63  & 0.57    & 0.56   & 0.49          & 0.52      & \textbf{0.46}  \\  
FID $\downarrow$ & 19.68  & 18.74  & 19.34    & 18.97   & 19.85          & 20.07      & \textbf{18.46}  \\\bottomrule
\end{tabular}
\end{table*}

\begin{table*}[t!]
\centering
\small
\caption{The average inference time (T) and GPU memory consumption (M) on four image resolutions. ``OOM" means ``cuda out of memory". }
\renewcommand{\arraystretch}{1.1}
\setlength{\tabcolsep}{3pt}
\label{time_com}
\begin{tabular}{c|cccccccccccccc}
\hline
  \multicolumn{1}{c|}{\multirow{2}{*}{\textbf{Resolution}}}& 
  \multicolumn{2}{c}{\textbf{AdaIN} \cite{adain}} &
  \multicolumn{2}{c}{\textbf{SANet} \cite{sanet}} &
  \multicolumn{2}{c}{\textbf{AdaAttN} \cite{adaattn}} &
  \multicolumn{2}{c}{\textbf{StyTr$^2$} \cite{stytr2}} &
  \multicolumn{2}{c}{\textbf{StyA2K} \cite{stya2k}} &
  \multicolumn{2}{c}{\textbf{AesPA-Net} \cite{aespa-net}} &
  \multicolumn{2}{c}{\textbf{Ours}} \\ 
     & \textbf{T(s)} & \textbf{M(GB)} & \textbf{T(s)} & \textbf{M(GB)} & \textbf{T(s)} & \textbf{M(GB)} & \textbf{T(s)} & \textbf{M(GB)} & \textbf{T(s)} & \textbf{M(GB)} & \textbf{T(s)} & \textbf{M(GB)} & \textbf{T(s)} & \textbf{M(GB)} \\ \hline
$256 \times 256$  & 0.0029  & 0.13       & 0.0038  & 0.17       & 0.0038  & 0.21       & 0.0255  & 0.41       & 0.0036  & 0.21       & 0.0057  & 0.67       & 0.0076  & 0.23       \\
$512 \times 512$  & 0.0030  & 0.45       & 0.0048  & 0.48       & 0.0487  & 1.32       & 0.1326  & 1.67       & 0.0038  & 1.13       & 0.0503  & 2.40       & 0.0081  & 0.61      \\
$1024 \times 1024$ & 0.0035  & 1.71       & 0.0051  & 2.48       & 0.1538  & 7.31       & OOM     & OOM        & 0.0053  & 6.01       & 0.2235  & 10.06      & 0.0203  & 2.11       \\
$2048 \times 2048$ & 0.0092  & 6.75       & OOM     & OOM        & OOM     & OOM        & OOM     & OOM        & OOM     & OOM        & OOM     & OOM        & 0.1535  & 8.12       \\ \hline
\end{tabular}
\vspace{-0.5cm}
\end{table*}

\section{Experiments}
\subsection{Implement Details}
We train our model using the MS-COCO dataset \cite{mscoco}, which contains approximately 120k real photos, and the WikiArt dataset \cite{wikiart}, which includes around 80k artistic images. During the training phase, we first resize each training image to 512 $\times$ 512 and then randomly crop it to 256 $\times$ 256 as the input. We utilize Adam as the optimizer. The batch size and learning rate are set as 4 and 0.0001 respectively. We train our model for 100k iterations on an NVIDIA GTX 4090Ti for about four hours.

\subsection{Comparison with State-of-the-Art Methods}
To evaluate the performance of our method, we compare it with six state-of-the-art AST methods: AdaIN\cite{adain}, SANet\cite{sanet}, AdaAttN\cite{adaattn}, StyTr$^2$\cite{stytr2}, StyA2K\cite{stya2k} and AesPA-Net\cite{aespa-net}. Among these, SANet and AdaAttN are the pure attention based methods. We also employ some improved attention based approaches like the StyA2K and AesPA-Net to compare stylization performance. All results of these methods are obtained by running their public released codes with default settings.

\textbf{Qualitative Evaluation.} 
The comparison in Figure \ref{exp_results} demonstrates the superiority of the proposed method in visual quality. Our stylized results faithfully reflect the style elements such as colors and strokes from the style image, and no local collapse in the stylization results. Additionally, our results maintain an excellent consistency in content structure compared to other methods.
In contrast, the generated images of AdaIN \cite{adain} lack sharp details and fine brushstrokes because its transfer process does not combine style semantic information. 
Since the style transformation of SANet \cite{sanet} and AdaAttN \cite{adaattn} is entirely based on the attention mechanism, they may repeat local salient patterns of style image, such as the repetitive eyes of human faces in the 1st row and the clouds in the 2nd row. StyTr$^2$ \cite{stytr2} and StyA2K \cite{stya2k} tend to preserve the content structure excessively, which easily results in an under-stylization, such as the face and collar of the man in the 1st row still remain some colors of content image. AesPA-Net \cite{aespa-net} can reduce the pattern repetition caused by the attention module, but it also inadvertently erases the brushstrokes that increase the beauty of the stylized images. In general, our method achieves a better balance in content preservation, stylization depth and visual quality compared to the comparison methods.

\textbf{Quantitative Evaluation.}
We use content loss, style loss, LPIPS \cite{lpips} and FID \cite{fid} to evaluate the visual quality of stylized images. Content/style loss measures the content/style consistency between the content/style image and the stylized images. LPIPS and FID evaluate the content/style fidelity between the stylized image and the corresponding content/style image.
For these metrics, a lower score indicates better quality of the generated stylizations. We randomly selected 20 content and 20 style images to generate 400 stylized images for comparison. Table \ref{fid} shows the experimental results.
As we can observe, our method achieves the lowest style loss, LPIPS and FID, and is only slightly higher than StyA2K in content loss, which shows that our method is significantly competitive in both transferring style information and preserving content structure compared with the state-of-the-art methods.

\textbf{Efficiency Analysis.}
In Table \ref{time_com}, we compare the inference time and GPU memory consumption at four image resolutions: 256 $\times$ 256, 512 $\times$ 512, 1024 $\times$ 1024 (1K) and 2048 $\times$ 2048 (2K). 
All experiments are run on an NVIDIA 4090Ti GPU. 
We can observe that the models containing the attention module all face memory overflow problems when processing 2K resolution.
The most obvious is StyTr$^2$ that based on Transformer architecture, which start overflows from 1K resolution. This is because the computational complexity of these models is quadratic with the size of the input image. In contrast, our HSI module achieves linear complexity with the image size. The results show that our method can run successfully at any test resolution. When processing 1K images, our method achieves about 50 fps and occupies only about 2GB of GPU memory.  The results show that our method is capable of performing style transfer in real time.

\begin{figure} [t!]
\centering
\setlength{\abovecaptionskip}{-0.1cm}
\includegraphics[width=3.5in, height=2.6in]{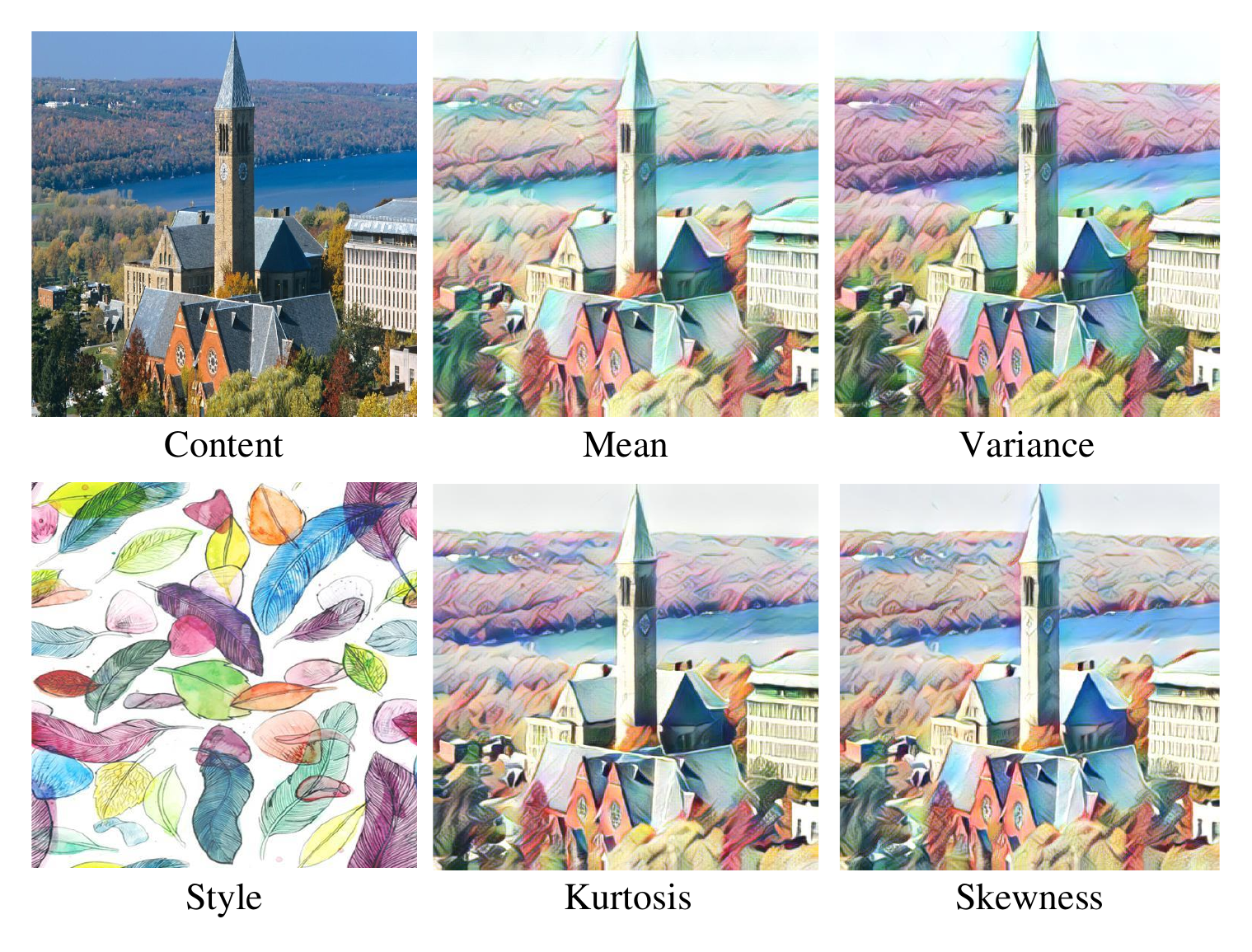}
\caption{Effectiveness of four style statistic features.}
\label{four_statistics}
\vspace{-0.5cm}
\end{figure}

\begin{figure} [t!]
\centering
\setlength{\abovecaptionskip}{-0.1cm}
\includegraphics[width=3.5in, height=1.2in]{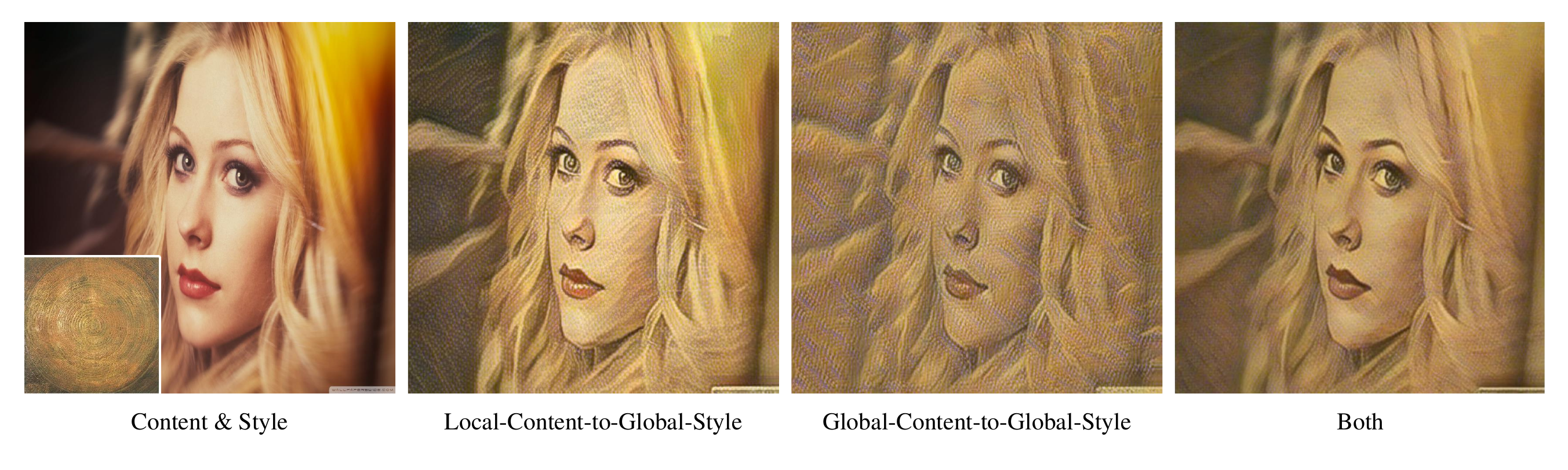}
\caption{Effectiveness of dual relations construction.}
\label{dual_relation}
\vspace{-0.5cm}
\end{figure}

\begin{figure*} [t!]
\centering
\setlength{\abovecaptionskip}{-0.1cm}
\includegraphics[width=6.0in, height=2.6in]{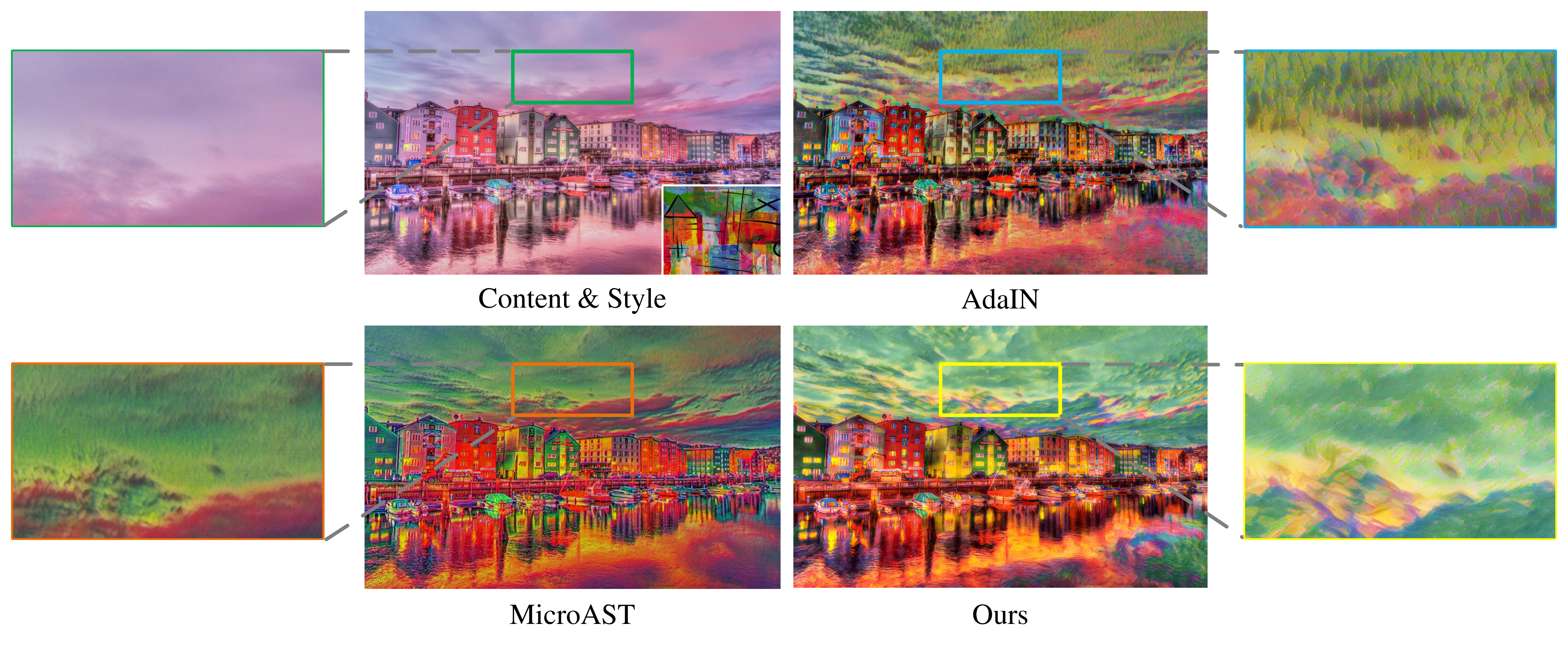}
\caption{Comparison on high-resolution (2K) style transfer.}
\label{high_resolution}
\vspace{-0.4cm}
\end{figure*}

\begin{figure} [t!]
\centering
\setlength{\abovecaptionskip}{-0.1cm}
\includegraphics[width=3.5in, height=1.8in]{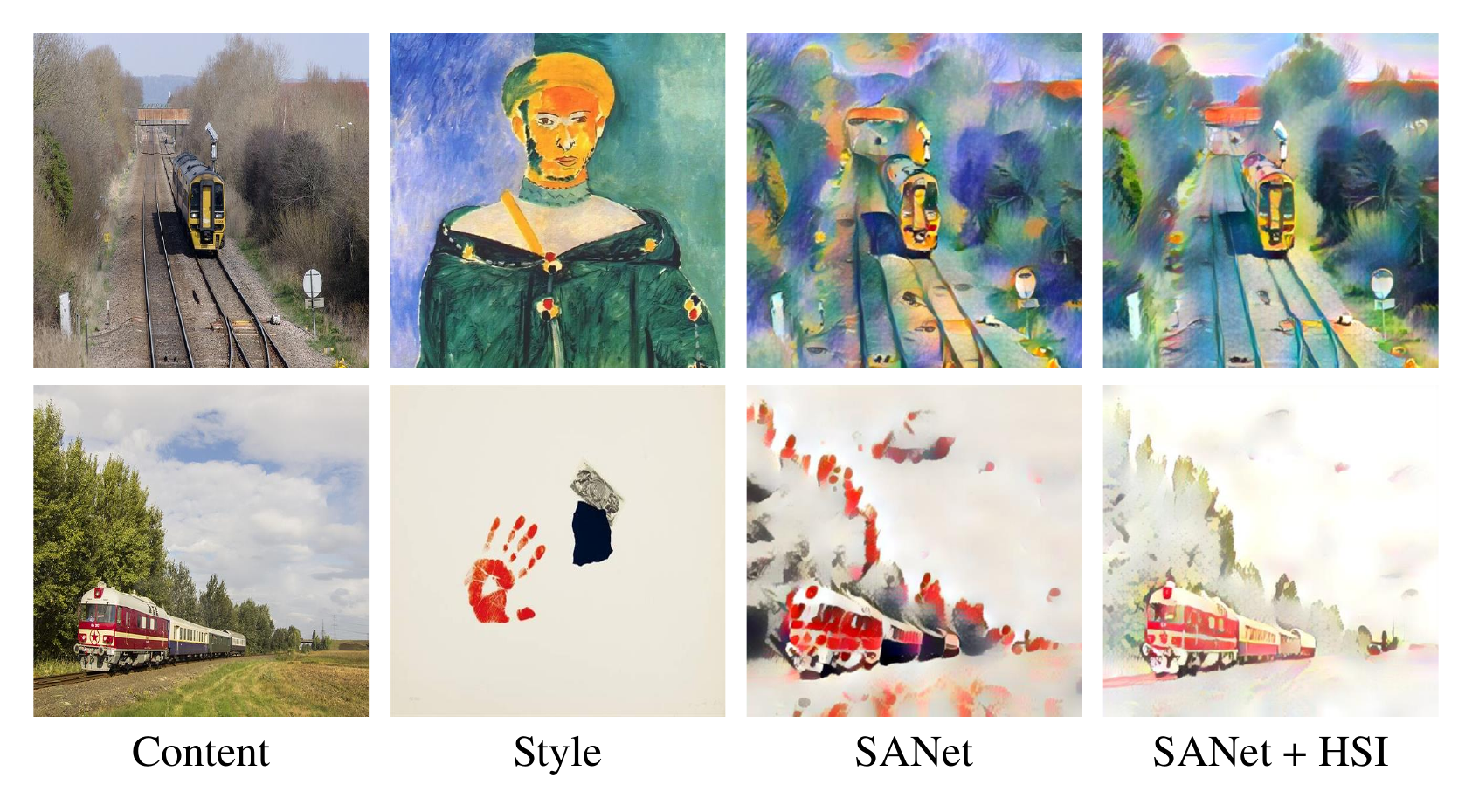}
\caption{Generalization of HSI module. By replacing the self-attention in SANet with our HSI module, the distorted regions and disharmonious patterns in the results of original SANet are reduced significantly, and the semantic structure of content image is well preserved.}
\label{generalization}
\vspace{-0.5cm}
\end{figure}

\subsection{Ablation Study}

\textbf{Effectiveness of Four Style Statistic Features.} In order to verify the effectiveness of four style statistical features (mean, variance, kurtosis and skewness), we interact each style features with the content features separately. As shown in Figure \ref{four_statistics}, each statistical feature can bring different colors and stroke in different content areas. This ensures that HSI can inject richer style elements into the stylized image and has more possibilities to improve the visual effect of the resulting image.

\textbf{Effectiveness of Dual Relations Construction.} To verify the effectiveness of the dual relations construction mechanism, we construct only local-content-to-global-style or global-content-to-global-style relation and both in HSI module. As shown in Figure \ref{dual_relation}, we can observe that the stylized image preserves more content details when only constructing a local-content-to-global-style relation, while constructing global-content-to-global-style relation can better integrate the global style. 
The resulting image brought by dynamically combining both relations can fuse their respective advantages. This proves that constructing two relationships at the same time can improve the consistency of content and style of the resulting image.

\subsection{More Discussions}

\textbf{High-Resolution Image Style Transfer.}
To verify the stylization effectiveness of our method on high-resolution images, we conducted experiments using 2K resolution images. We compared our proposed method with AdaIN and MicroAST \cite{microast}. MicroAST is a professional high-resolution style transfer model. The experiment results are shown in Figure \ref{high_resolution}. Our method renders style elements such as brushes and colors more abundantly and compared with AdaIN and MicroAST, which shows its effectiveness in transferring holistic style patterns.

\textbf{Generalization of HSI module.}
The HSI module proposed in this paper is a plug-and-play module. To verify its generalization performance, we replace the attention module in SANet with the HSI module, denoted as SANet + HSI. We retrain the model with the default settings and compare it with the original SANet. The results are shown in Figure \ref{generalization}. In the results of the original SANet, we can observe obvious overly repetitive style patterns, such as the faces in the 1st row and the red palm prints in the 2nd row, which cause local distortions that greatly reduce the visual quality. In contrast, when the HSI module replaces the attention module, the repeated patterns and distorted areas in the stylized image are significantly reduced. This further demonstrates the effectiveness of the HSI module in both capturing the overall style pattern and preserving the content structure.

\section{Conclusion}
In this paper, we propose a new attention-style transfer module, named Holistic Style Injector (HSI), for efficient and effective arbitrary style transfer. It contains three characteristics: Global styles extraction, dynamic dual relations construction, and the linear transfer process. Global style extraction emphasizes rendering various global style patterns on the content image while preventing local distortions and enriching style information. 
Dynamic dual relation construction establishes local-content-to-global-style and global-content-to-global-style relations simultaneously. It flexibly adjusts the matching between style and content distribution according to semantic similarity to obtain more harmonious stylization results. 
The linear transfer process refers to the style integration in our model based on efficient element-wise multiplication, which can complete stylization in a linear complexity. 
Extensive experiments demonstrate the superiority of our method.

{\small
\bibliographystyle{ieee_fullname}
\bibliography{egbib}
}

\end{document}